# Rethinking the Physical Symbol Systems Hypothesis

Paul S. Rosenbloom

Department of Computer Science
University of Southern California, Los Angeles CA 90089, USA
`rosenbloom@usc.edu`

**Abstract.** It is now more than a half-century since the *Physical Symbol Systems Hypothesis* (*PSSH*) was first articulated as an empirical hypothesis. More recent evidence from work with neural networks and cognitive architectures has weakened it, but it has not yet been replaced in any satisfactory manner. Based on a rethinking of the nature of computational symbols – as *atoms* or *placeholders* – and thus also of the systems in which they participate, a hybrid approach is introduced that responds to these challenges while also helping to bridge the gap between symbolic and neural approaches, resulting in two new hypotheses, one that is to replace the PSSH and other focused more directly on cognitive architectures.

## 1   Introduction

Our current understanding of the role of *physical symbol systems* in artificial intelligence (AI) is grounded in the pioneering work of Newell and Simon [1-3], although as they point out the roots go back much further in philosophy – most notably in logic – computer science, linguistics, literature, and the arts. Such systems, and their culmination in the *Physical Symbol Systems Hypothesis* (*PSSH*) are reviewed in Section 2.

Many critiques of the PSSH have been proposed since it was first introduced, with some that have easily been refuted and others that have lingered (Section 3). Here, two are taken up that have remained compelling, before *hybrid symbol systems* of a particular sort are explored as a response to them (Section 4). As part of this, the notion of symbol systems is rethought, starting with a variant definition of what it means to be a computational symbol that is grounded in the Common Model of Cognition (CMC) [4] and the Sigma cognitive architecture [5]. Two new hybrid hypotheses result, one that offers an alternative to the PSSH and the other that focuses more specifically on cognitive architectures.

Demonstrating that neural networks are themselves hybrid symbol systems of this sort (Section 5), rather than being limited to the numeric component of a coarse-grained combination of symbolic and numeric processing, helps to bridge the gap between symbolic and neural approaches while enabling recent successes with neural networks to be weighed in a positive manner in evaluating hypotheses concerning symbol systems, rather than the former necessarily serving as a challenge to the latter.

The overall result, as discussed further in Section 6, is a novel way of thinking about symbol systems and the fundamental hypotheses concerning them; the introduction of a particular form of hybrid symbol system and the appropriate hypotheses concerning it; and an understanding of how neural networks are examples, rather than counterexamples, of this form of symbol system. The hope is that this all helps cut the Gordian Knot that has resulted from past discussions on these topics.

Proposing hybrid or neuro-symbolic systems is certainly nothing new. Many approaches have already been investigated – see, e.g., [6] and [7] for overviews, and [8] for an earlier discussion of the PSSH and the relevance of hybrid systems. But the point here is to introduce a particular take on hybrid symbol systems that is in service of an appropriate rethinking of the Physical Symbol Systems Hypothesis. The approach is broader than neuro-symbolic, as it also includes hybrid systems that span other numeric paradigms, such as probabilities. In addition, it spans both tightly coupled and loosely coupled approaches to combining symbolic and numeric processing.

## 2  Physical Symbol Systems

According to the traditional view, *symbols* are distinct patterns in the physical world that can be *composed* into *expressions*, or *symbol structures*. *Processes* are then defined on these symbol structures that can create, modify, reproduce, and destroy them. An expression *designates* an entity, whether internal or external, if the expression's use depends on the nature of the entity. An expression is *interpreted* if it designates an internal procedure that is then executed. The physicality of such symbol systems reflects that they are *natural*, in obeying the laws of physics and being amenable to engineering; and that they aren't limited to what is in human minds, or even necessarily based on the same kinds of symbols that have traditionally been imputed to humans.

Given composition, designation, and interpretation, along with the appropriate processes, physical symbol systems provide a form of universal computation. There are certainly more details in the various papers, but this provides the essence of what can now be considered the classical notion of a physical symbol system.

The *Physical Symbol Systems Hypothesis* (PSSH) then states that:

> A physical symbol system has the necessary and sufficient means for general intelligent action.

This hypothesis was introduced as an empirical generalization rather than a theorem. Evidence for sufficiency stemmed from the universality of symbol systems and the success of such systems built as of then. Evidence for necessity stemmed from noting that the one natural system exhibiting such intelligent behavior – that is, humans – appeared to be such a system, and from the lack of alternative approaches that were nearly as successful. Newell, for example mentions that "These advances far outstrip what has been accomplished by other attempts to build intelligent mechanisms, such as the work in building robots driven directly by circuits; the work in neural nets, or the engineering attempts at pattern recognition using direct circuitry and analogue computation." [3].

He went on to state that "In my own view this hypothesis sets the terms on which we search for a scientific theory of mind." and "The physical symbol system is to our enterprise what the theory of evolution is to all biology, the cell doctrine to cellular biology, the notion of germs to the scientific concept of disease, the notion of tectonic plates to structural geology."

## 3  Critiquing the Physical Symbol Systems Hypothesis

It has now been over fifty years since the PSSH was first articulated, with numerous critiques and defenses occurring in the intervening years. Nilsson [8], e.g., lists four general types of critiques with his responses to them (in italics here), which in brief are:

1. Lack of embodiment/grounding.
   *This is a misunderstanding as the PSSH already includes this.*

2. Non-symbolic/analog processing.
   *Include numbers; that is, make the systems hybrid.*

3. Brain-style versus computation-style (i.e., brains are not computers).
   *The brain is computational.*

4. The mindlessness of much of what appears to be intelligent behavior.
   *Mindless constructs only yield mindless behavior.*

In this section two particular critiques are considered, based on new empirical evidence in the form of the recent successes of deep learning [9], and to a lesser extent probabilistic graphical models (PGMs) [10], plus work on the CMC. One critique, aligned with Nilsson's second, challenges its sufficiency and the other its necessity.

The sufficiency challenge focuses on the lack of numeric processing – i.e., calculations on quantities – in the PSSH. Nilsson's response is to shift to hybrid systems that include both symbols and numbers. In a sense, this isn't logically necessary, as the universality of symbol systems implies that, as with any modern digital computer, they can implement algorithms for numeric processing. However, universality is weaker than what was originally proposed, as it omits grounding sufficiency in the successes of existing symbolic AI systems. Given the range of general intelligent action that has been shown to proceed more effectively with numeric processing, whether in the form of probabilities or activations, the success of purely symbolic systems no longer provides compelling empirical evidence itself for the sufficiency of symbols on their own.

Thus, we are left with a weakened form of sufficiency for the PSSH, based solely on universality. Hybrid systems have the potential to restore the stronger sense of sufficiency (Section 4). They also support a more stringent sufficiency hypothesis that arises when the concern is more particularly with cognitive architectures [11]; that is, models of the fixed structures and processes that yield a mind [12].

The necessity challenge is rooted directly in how neural networks now provide a better approach for many problems related to intelligent action. Successes with PGMs can be considered here as well, although they are already hybrid systems that add probabilities to classical symbol systems, particularly in their most general form as statistical relational systems [13], so they do not directly challenge the necessity of physical symbol systems. In contrast, deep learning has the potential to provide an alternative that completely overturns the necessity argument. In Section 5, this challenge is approached via a demonstration that, given the rethinking of symbol systems in Section 4, neural networks are themselves instances of hybrid symbol systems. This approach avoids the need to resolve the contentious question of whether or not neural networks have or need traditional symbols, a question that appears unresolvable, at least to me, without additional evidence.

## 4 Rethinking Symbol Systems

This section leverages the four-step methodology of *essential analysis* [14] to yield a fresh understanding of symbols and symbol systems: (1) strip out many of the elaborations that are normally part of a topic's definition, and which are often a source of dissonance among researchers and communities, to yield its *essence*; (2) use what has been stripped out, and possibly more, in specifying a *definitional space* of variations on the topic; (3) populate this space with *exemplars* that flesh it out; and (4) derive novel *implications* from the results of the first three steps. Step three is downplayed here due to lack of space, while step four introduces two new hybrid symbol systems hypotheses.

The focus here is in particular on the notion of symbol as it is used computationally rather than as it is used in the humanities and arts. For example, [15] defines a symbol as "something used for or regarded as representing something else; a material object representing something, often something immaterial; emblem, token, or sign." This focuses on an abstract notion of *designation* or *aboutness*, which has elsewhere been considered an important part of the essence of a theory [14]. Computationally, the essence of a symbol is proposed to be an *atom* that is: (1) *indecomposable* into other atoms; and (2) *distinct* from other atoms. McDermott informally introduced the notion of a symbol as a *placeholder* [16]. Although yielding different connotations, this notion is compatible with that of an atom here.

This essence retains the classical notion of a computational symbol being a primitive element that can be distinguished from other such elements but eschews the need for both physicality and symbols being structured as patterns. There were good reasons at the time to emphasize physicality – to counter both Cartesian dualism and the notion that only humans could use symbols – but these battles have already been won, at least in my judgement, so this explicit emphasis on physicality is now dispensable.

Pattern comparison is one way to determine whether two atoms are distinct. Yet, such a notion need not be definitional if it is just used to compare symbols. If symbols are considered as *types* (rather than *tokens*) – a notion implicit in the traditional definition – patterns are simply *intensional* definitions of symbols. An *extensional* alternative defines each symbol in terms of a set of tokens, with each token in a set considered to be indistinct from other tokens in the same set and distinct from tokens in other sets.

The classical notion of symbol also includes *composability* – into symbol structures or expressions – *designation*, and *interpretation*. The first of these is effectively assumed to be part of the very nature of symbols, whereas the latter two are additional properties necessary to enable the classical form of physical symbol systems. The essential definition of a symbol introduced here includes none of these three notions; that is, all are optional. Therefore, any system that includes even these minimal, atomic forms of symbols can be considered a symbol system of some sort.

Fig. 1 structures these optional properties, plus a bit more, into a small tree. According to this perspective, a symbol may be *composable* into expressions (aka symbol structures). It may also *designate*; that is, stand in for something else. A designation is *procedural* if it is about a process. This is the classical notion of interpretation, when combined with the ability to execute the designated process. A procedural symbol, according to this definition, designates a process rather than being part of the process itself. If the process is itself a symbol structure it will contain symbols, but they themselves may be of any type.

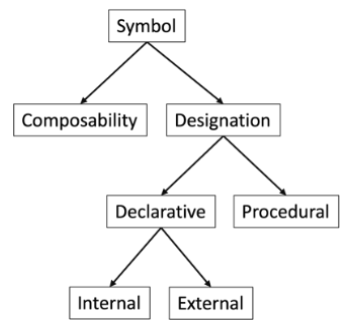

**Fig. 1.** Optional properties of symbols.

A designation is *declarative* if it is about an object – essentially anything other than a process – which may be *internal* to the system or *external* to it, with the latter relating to grounding. This corresponds to the classical notion of designation when contrasted with interpretation. Beyond this difference in what is designated, there is no intent to impute any other aspects of the classical procedural versus declarative distinction here.

Symbols in a classical symbol system support all of these properties, enabling them to exhibit computational universality. Whether systems in which some or all of the symbols lack some of these properties provide anything like universal computation would necessarily depend on the details of the individual systems.

The CMC, an attempt at developing a consensus on what is needed for human-like cognition – i.e., human cognition and similar forms of artificial cognition – took a step towards such an essence by dropping the necessity of designation, and thus also of interpretation, stripping symbols down to primitive elements that only support composability into symbol structures. Although designation somewhere in a system seems necessary for it to be either meaningful or operational, it is not necessary for all symbols.

The CMC also associated *quantitative metadata* with such symbols and structures – which provide the *data* – to modulate how they are processed. Such combinations can be considered as *hybrid symbols* or *structures*.[1] Considering *hybrid* of this sort as a third optional property of symbols leads to Fig. 2.

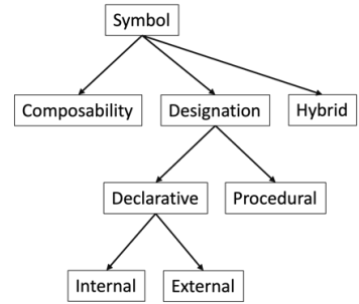

**Fig. 2.** Optional properties of symbols from Fig. 1, extended with hybrid.

The CMC went on to argue for a different form of weakening of the sufficiency aspect of the PSSH. While still agreeing that classical symbol systems, as universal computational systems, are sufficient in principle for intelligent behavior, it denied that they are sufficient when time scales are relevant, such as in cognitive architectures. In particular, if statistical processing must occur on the same time scale as symbolic processing in such an architecture, then implementing the former in terms of the latter – as the universality argument for sufficiency implies – is insufficient. Thus, the CMC implies the need for numeric and symbolic processing on the same time scale. The Sigma cognitive architecture [5] goes a step further by denying the need for all symbols to support arbitrary forms of composition, thus implicitly yielding the essence made explicit here.

Now, given this explicit articulation of the essence of a symbol plus its tree of variations, the *Hybrid Symbol Systems Hypothesis* (*HSSH*) can be stated as:

> Hybrid symbol systems are necessary and sufficient for general intelligent action.

If the sufficiency clause of the PSSH is valid then so must be the comparable clause in the HSSH, at least for hybrid symbol systems that are universal. However, the HSSH responds to the PSSH sufficiency challenge by including numbers, as suggested in [8]. Necessity of the HSSH is not implied by the corresponding clause in the PSSH. Instead, the HSSH responds to the PSSH necessity challenge by coopting the successes of neural networks (Section 5).

The *Hybrid Cognitive Architectures Hypothesis* (*HCAH*) then states:

> Hybrid symbol systems are necessary and sufficient for cognitive architectures.

This hypothesis is clearly related to the HSSH, but it matters in itself because the comparable hypothesis – perhaps called the *Physical Cognitive Architectures Hypothesis* (*PCAH*) – fails. Thus, the sufficiency side of the PCAH is invalid irrespective of what might be true with respect to necessity. As with the HSSH, sufficiency for the HCAH need not hold for all hybrid symbol systems, but it must hold for at least some.

As with the PSSH and the HSSH, the HCAH is an empirical generalization. Both sides of the argument are now supported by the architectural successes of classical symbol systems, neural systems, and traditional hybrid systems such as PGMs. Both sides are further bolstered by how the CMC itself is a hybrid symbol system.

---

[1] The CMC also allows numeric data, consideration of which is beyond the scope of this paper.

## 5     Neural Networks as Hybrid Symbol Systems

What makes neural networks hybrid symbol systems, as defined here, rather than simply the numeric component of a larger system that also includes a symbolic component, such as [17]? To keep things simple, the focus here is limited to standard feedforward neural networks, consisting of multiple layers of nodes and links, where nodes have activations, links connect pairs of nodes across levels and have weights, and processing occurs by multiplying input activations along links by the links' weights and then nonlinearly transforming the sums of these weighted inputs.

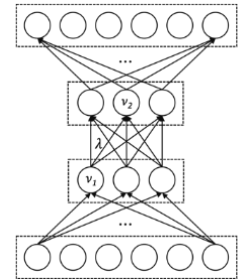

**Fig. 3.** Simple network for paired associates.

To be a bit more specific, let's assume a small network for *paired associates* that maps an input word to an output word. Fig. 3 exemplifies this via a completely connected network with 6-unit input and output layers – yielding a 6-dimensional vector of activations for each – and 2 intermediate layers, each with 3 units. Words map consistently to input and output vectors via *encoding* and *decoding* processes that are external to the network. These processes may be based on an arbitrary or random assignment of vectors to words or some form of more sophisticated *embedding* process, such as in [18].

The focus here, however, is on analyzing the forward processing in the network itself to show how it amounts to a hybrid symbol system. It should be possible to extend such an analysis to encoding and decoding processes, as well as to learning in neural networks, but this simple example is sufficient to establish the precedent.

First consider the nodes in the input layer of the network, now shown at the top of Fig. 4 as locations within a vector of nominal activations. Such nodes can be seen as hybrid symbols – symbolic nodes (i.e., locations) with activations as their quantitative metadata – that exhibit a limited form of composability in yielding the hybrid symbol structure that is the input vector. In contrast to the traditional interpretation of distributed representations – where nodes are subsymbolic, or microfeatures, with symbols only arising as patterns over these elements – here the individual nodes are themselves hybrid symbols that do not themselves designate, with patterns arising as structures of these hybrid symbols.

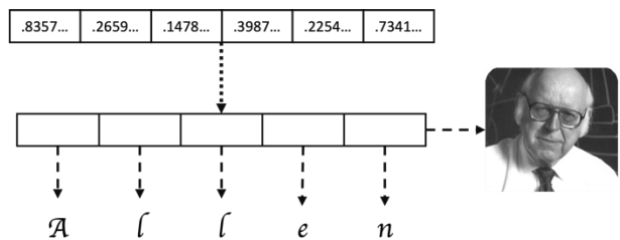

**Fig. 4.** Designation relationships among input vector, word vector, letters, and word meaning. Dotted lines reflect internal designation and dashed lines external designation.

This hybrid symbol structure does then internally designate a word structure that has one location per letter (middle of Fig. 4). The metadata in the word structure is not shown as it is irrelevant to this analysis. What does matter is that declarative symbols in this word structure externally designate particular letters of the alphabet, in this case making up Allen Newell's first name (bottom of Fig. 4). The word as whole then externally designates its meaning, iconified to the right of Fig. 4 via an image of him.

Key to this all working is that it isn't just the data aspect of hybrid symbols and structures that can designate, but the entirety of the hybrid symbols and structures – including their metadata – that can do so, just as is traditionally assumed for vectors in distributed representations [19]. The word itself is epiphenomenal to the feedforward network processing here – only the hybrid symbol structure at the top of Fig. 4, as yielded by encoding, actually participates. As put in [20], "the node labels in a Connectionist machine are not part of the causal structure of the machine."

The internal nodes in the network are also hybrid symbols but without declarative designations, fitting the intuition that there are no fixed meanings inside the network. Instead, internal nodes – and links – procedurally designate fixed processes. Consider link $\lambda$ in Fig. 3, which points from node $v_1$ to node $v_2$. This link is a hybrid symbol structure composed from these two hybrid symbols, with a weight as its metadata. It procedurally designates a process that multiplies the activation arriving from $v_1$ by this weight. Internal nodes such as $v_2$ then procedurally designate processes that sum all of their inputs – in this case, from $v_1$ and any other nodes linked to it from the proceeding layer – and then nonlinearly transform the results.

The last part of the analysis concerns the output nodes. Perhaps surprisingly, they too do not declaratively designate anything here. Instead, they procedurally designate the same summation and transformation process as the internal nodes. It is not until postprocessing – that is during decoding – that this reverse mapping occurs.

This analysis demonstrates that a feedforward neural network is a hybrid symbol system, as defined here. As such, it makes the case that the shift from the PSSH to the HSSH enables coopting neural-network successes as evidence for both the sufficiency and necessity of hybrid symbol systems rather than as counterexamples to them.

But what type of hybrid symbol system does this type of neural network yield? It provides limited forms of declarative designation (at input nodes), procedural designation (at all but input nodes), and composition (via vectors within a level and links across levels). Yet, other forms of neural networks do go beyond this. To name just two common examples, both convolutional networks (e.g., [21]) and transformers [22] include additional forms of composition. The flexibility of composition seen in the output of transformer-based generative networks [23] is in fact quite compelling. Some forms of neural networks are also known to support universal computation (see, e.g., [24]). Yet no neural network to date has solved combinatorial board games without the dynamic composition yielded by explicit state-space search (as seen, e.g., in both [25] and [26]). So, the overall story is complex, dependent on the exact nature of the neural networks considered, and still not completely understood.

## 6  Conclusion

Leveraging essential analysis, symbols are (re)defined as atoms or placeholders, and a space of variations is defined for symbols, symbol structures, and symbol systems. This includes the classical traits of compositionality and designation, plus hybridness and additional sub-traits under designation (such as interpretation). In response to lingering challenges to the Physical Symbol System Hypothesis (PSSH), two new hypotheses have then been introduced that focus on the resulting hybrid symbol systems:

Hybrid Symbol Systems Hypothesis (HSSH):

> Hybrid symbol systems are necessary and sufficient for general intelligent action.

Hybrid Cognitive Architectures Hypothesis (HCAH):

> Hybrid symbol systems are necessary and sufficient for cognitive architectures.

The HSSH is intended as a replacement for the PSSH, based on evidence accumulated since the latter was introduced as an empirical hypothesis a half-century ago. Given this recent body of evidence, there is a sense in which the PSSH still holds, but it is a weaker sense. The HSSH recaptures the originally intended strength while adding further to it by reinterpreting neural networks as compatriots – that is, as hybrid symbol systems themselves – rather than as competitors. The result also helps chip away in a rather fine-grained manner at the overall divide between symbolic and neural systems.

The HCAH is a more stringent claim than either the original PSSH or the HSSH in that it concerns cognitive architectures rather than general intelligent action. Evidence accumulated over the past decades has shown that traditional physical symbol systems fail with respect to sufficiency for cognitive architectures due to the need for numeric processing within the architectures themselves. The necessity of classical physical symbol systems for cognitive architectures remains an open question, as it is not yet clear whether neural networks – which although as argued here are hybrid symbol systems but which may not be classical symbol systems or even universal computationally – will prove to be a sufficient alternative on their own for such architectures.

One potential chink in the armor of both of these new hypotheses is the possibility of quantum aspects to intelligence that cannot be captured even by hybrid systems [27]. Should it prove necessary, some thought is already being put into what it would mean to have quantum symbol systems (e.g., [28]).

## Acknowledgements

I would like to think John Laird, Christian Lebiere, and Andrea Stocco for helpful comments and discussions on this general topic and this particular paper.